# Robot Affect: the Amygdala as Bloch Sphere

Johan F. Hoorn[1,2]

Johnny K. W. Ho[3]

[1] Department of Computing and School of Design, The Hong Kong Polytechnic University, 999077, Hong Kong.
[2] Department of Communication Science, Vrije Universiteit Amsterdam, 1081 HV, The Netherlands.
[3] Department of Physics, Hong Kong Baptist University, 224 Waterloo Road, Kowloon Tong, Kowloon, Hong Kong SAR.

Correspondence should be addressed to Johan F. Hoorn; csjfhoorn@comp.polyu.edu.hk

## Motto

$(2x^2 + y^2 + z^2 - 1)^3 - (1/10) \cdot x^2z^3 - y^2z^3 = 0$ (Taubin, 1994, p. 22)

## Abstract

In the design of artificially sentient robots, an obstacle always has been that conventional computers cannot really process information in parallel, whereas the human affective system is capable of producing experiences of emotional concurrency (e.g., happy *and* sad). Another schism that has been in the way is the persistent Cartesian divide between cognition and affect, whereas people easily can reflect on their emotions or have feelings about a thought. As an essentially theoretical exercise, we posit that quantum physics at the basis of neurology explains observations in cognitive emotion psychology from the belief that the construct of reality is partially imagined (Im) in the complex coordinate space $\mathbb{C}^3$. We propose a quantum computational account to mixed states of reflection and affect, while transforming known psychological dimensions into the actual quantum dynamics of electromotive forces. As a precursor to actual simulations, we show examples of possible robot behaviors, using Einstein-Podolsky-Rosen circuits.

在机器人的情感拟人化设计中，一般电脑都没法做到并列信息处理，但人类的情感系统往往拥有同时产生多种感受的能力（如悲喜交集），这成为实现情感拟人化一直而来的一大障碍。另一大障碍是人的情感和认知的笛卡尔二元性，该二元性让人可以容易地对他们的情感加以思考，也可以对他们的想法产生感受。本论文从理论层面出发，以神经科学为基础的量子物理的角度，展释认知情绪心理学中的现象是来自一种原理，指出人对现实的认知是部分建基于一个复数空间($\mathbb{C}^3$) 里的虚部(Im)。我们提出一个由情感和思考组成的量子混合态的计算描述方式，并将已知的心理维度转换成实在的量子力学体现及其可量度的电动势。作为模拟计算的先导，我们透过爱因斯坦-波多尔斯基-罗森量子线路，展示一些机械人可能出现的行为。

# Introduction

To open up psychological processing to quantum computing, we developed an understanding of human information processing in terms of mixes of reflective and affective operations expressed as Bloch vectors. We conceive of information as oscillations of electrons that can be superposed, resulting into ‘mixed states’ of reflection and affect, which are described by the probability distributions of the multiple pure states that the oscillations can be in. Purely theoretically, this paper is a contribution to psychology in providing a radically physical and fully quantifiable model of the functional make-up of certain neural pathways. As an exercise, this paper will be useful to make future robots process emotional data and simulate cognitive-affective processes in a human-like fashion (cf. Raghuvanshi & Perkowski, 2010).

There already is a body of literature proposing that quantum mechanisms are active in human information processing. In their review, for example, Schwartz, Stapp, and Beauregard (2005) observed that modern physics takes into account psychological decisions in the explanation of causal physical relationships. Reversely, they observe that neuroscientists and psychologists increasingly (should) rely on quantum physics to describe neural processes that are determined by certain structural aspects of the ion channels that are operative in the synapses (the human information ‘switch boards’). These authors assert that “… contemporary physical theory must in principle be used when analysing human brain dynamics.”

Schwartz, Stapp, and Beauregard (2005) criticize contemporary brain science for assuming that measurable physiological data are the final explanation of psychological functions. Apart from the conundrum of ‘measurement’ in quantum physics (see the section entitled *Measurement: Imaginary, Real, and the Anger-and-Joystick*), these authors point out that contemporary neuropsychology cannot explain what happens *during* experimentation; how people may ‘willfully induce brain changes’ or employ ‘self-directed neuroplasticity,’ for instance, through training, cognitive reattribution, or conditioned attentional focus shifts (which may not be intended by the very experiment). Schwartz, Stapp, and Beauregard (2005) state that current neuropsychology should incorporate the mathematics of quantum physics to account for human observational bias in the measurement of physical properties of the human brain.

With its preoccupation of studying phenomena as discrete units, classic science struggles with the contextual aspects of an entity’s behavior and processes (whether in physics or psychology). Narens (2016, p. 323) indicates that psychology may find it difficult to include contextual aspects into its probabilistic models but that they can resolve this by applying quantum probability theory to handle the dynamics of contextual impact on behaviors.

In taking on the advice of Schwartz, Stapp, and Beauregard (2005) and Narens (2016), our main research question is this: If we should believe the neurologists and psychologists, then the firing frequencies of electrons carry information around the human brain. Firing frequencies translate to oscillations of electrons over a trajectory, say the ion shafts in the nervous system that allow neurons to generate action potentials. Is it so, then, that electrons, which through their oscillations carry information around the brain, are susceptible to basic quantum dynamics, including the superposition of an electron’s wave function over different locations in the brain - perhaps even other body parts, or other people? And if so, may superposed electrons explain the dynamics of human information processing, in particular of affect and reflection operating on the same piece of information in parallel?

The contents of this paper are as follows: With Raghuvanshi and Perkowski (2010) and Yan, Iliyasu, Liu, Salama, Dong, and Hirota (2015), we offer a Bloch sphere representation of psychological states but applied to the affective and reflective processing of information as described in LeDoux (1996/1999), Cerić (2012), and Crone and Konijn (2018). We then explore how psychological states can be modeled with the quantum logics of state vectors, while using Dirac (1958) notation. We proceed by giving psychological meaning to quantum-physical variables: Psychological relevance (Frijda, 2006) is the product of bigger or smaller bursts of energy (i.e. electric potential differences) in response to a stimulus. The experience of valence (ibid.) results from the position of electrons across the Amygdala and deep or shallow thinking pertains to the number of locations (Haller, 2016) the wave functions of electrons cover as ordained by the Neocortex.

Subsequently, we look at the external responses that should follow from our Bloch-sphere view and contemplate the use of quantum transition matrices and Einstein-Podolsky-Rosen circuits to allow for changes in entanglement of reasoning and affect. Applied to a model for a robot's attitude towards its user (Hoorn, Baier, Van Maanen, & Wester, submitted), we demonstrate the utility of state-transition networks for representing a dynamic cognitive-affective architecture and for discerning a system's potentially dangerous states.

Our final feat is the serious adoption of epistemological concerns (i.e. Hoorn, 2012) into the explanation of quantum events in the human brain. We show how measurement has a – to the observer's mental representation of the physical world – real (Re) aspect and simultaneously an imaginary side (Im and *i*), pertaining to the realm of $\mathbb{R}^3$ coordinate space and complex numbers in $\mathbb{C}^3$, respectively. We end on a daring proposal to - against better judgment - construct a drone-driven measuring instrument that navigates the Bloch sphere to test our quantum approach to psychological parallelism against real human beings. But first, we introduce some general principles of quantum physics.

## The Quantum Realm

A basic idea of quantum theory is that two aspects of the same entity that seem to exclude one another yet are concurrently present. For example, a particle such as an electron running through a brain circuit may be at different positions, at different levels of energy, flying at different speeds. Thus, a particle may not be in one state only but in an array of parallel states.

Merli, Missiroli, and Pozzi (1976) showed that a single electron could move around in the form of a wave, the one end of which was at one location, the other end at another location, both ends interfering with each other: Making each other stronger (constructive interference) (cf. standing sea waves) or canceling each other out (destructive interference) (cf. anti-sound).

Interestingly, if one attempts to observe all parallel states of a single particle such as an electron traversing a neuron, only one of those states will be measured (e.g., with electroencephalograms or EEGs). Superposition, however, says that one particle can be in two or more locations. Reversely, Pauli's exclusion principle forbids that two different electrons of the same spin can be in one location at once.[1] In other words, if a measurement device such as an electrode is brought close to an electron in superposition, that electron is forced by that device to evade the position the device occupies and so collapses back into one of two positions: The observation interfered with the object of study.

[1] Note that smaller particles such as bosons with integral spin do not obey Pauli's exclusion principle.

This observation problem is probably due to so-called ‘entanglement’ of quantum properties. When brought in each other’s vicinity, the particles of every object share or combine their properties or ‘features’ such that the observer cannot see the one entity without taking the other into consideration. Both entities start to behave as a unity, as one single entity that is composed of more, sometimes opposing, component parts. Through entanglement, objects start to pick up mass until at one point they are not susceptible to quantum rules alone anymore but enter the realm of classic mechanics as well. Physicist call the reverse transition (from classic to quantum world) ‘decoherence,’ which may not happen abruptly but rather gradually.

With respect to the superposition of oppositions, quantum theorists found a way to model the ambiguity of theoretically possible states as an alternative to common probability estimates. Owing to superposition, physicists defined the ‘quantum bit’ or ‘qubit’ by allowing in their theoretical models that a computer bit may not be either `0` or `1` (like an on-off switch on a conventional circuit board) but to be concomitantly `0` *and* `1`. When two particles become entangled, they share their quantum properties, including their respective superpositions. This means that the value of one qubit becomes dependent on that of the other.

Like this, one united superposition emerges from all possible combinations of values of the single qubits together. In the case of two qubits, four combinations are possible: `00`, `01`, `10`, `11`. Because the qubits are in superposition, the processing of one qubit also affects the processing of the other qubit, thus creating a parallel processor that codes for many combinations of `0`s and `1`s at once.

Conventionally, microchips carry many tiny capacitors, which are either charged or not charged. A standard hard drive represents information by running a current through numerous small electromagnets, magnetizing each either North of South. In a qubit, oppositions such as charged and not charged, North and South are available in unison.[2] Thus, where radio tubes, transistors, and microprocessors work in series of information that may run alongside one another, qubits truly work in parallel: The same information may be in multiple states.

In mathematical languages, a *complete set* of *orthogonal* quantities in the Hilbert space of an entity may serve as a qubit.[3] Note that orthogonal or ‘opposite’ quantities should be taken in a mathematical sense. In quantum processes, the qubit digits are expected to work in parallel. It is so not that coexistence leads to annihilation, for example, by combining opposing charges. While orthogonal quantities could be interpreted physically as oppositions, in terms of mathematics, they do not possess properties of reinforcement and cancellation by algebraic operations such as addition and subtraction. Different from classical mechanics, an electron could have a half-half probability of spin up and down, which perceptually may be *opposite* spins. Yet, such coexisting oppositions do not result into (algebraic) cancellation. Instead, they are two *orthogonal* qualities that allow to coexist.

---

[2] Some may say this is perceptual rather than actual.

[3] Hilbert space is Euclidian space but with an unlimited number of dimensions. Hilbert space thus encompasses Euclidian space.

In computation, the `0` and `1` in a particular binary digit may be regarded as opposite qualities (which we often refer to as up/down or on/off). In quantum computation, however, the basic units are $|0\rangle$ = (`1 0`) and $|1\rangle$ = (`0 1`) (see the section named *A Bloch Sphere of Emotions*). Now the basics are not numbers (scalars) but vectors, or more precisely, they are vectors of opposite physical interpretations but with orthogonal directions. Mathematically, one cannot find a combination of real numbers $a$ and $b$ such that $a\ |0\rangle + b\ |1\rangle$ = (`0 0`) except for the trivial solution $a = b = 0$. Therefore, $|0\rangle$ and $|1\rangle$ cannot cancel each other out, i.e. they are not opposite quantities that can algebraically undo each other but they exist in parallel as so-called quantum states.[4]

In our modeling (see *Psychological State Vector*), $|0\rangle$ represents the affective process (the Thalamus-Amygdala pathway) and $|1\rangle$ the reflective process (pathway via the Neocortex). While a contrasting perception can be inferred (fast/slow), the two pathways cannot cancel each other out. More specifically, pure states of greater weight in one pathway do not alter the length of the state vector on the Bloch sphere, but its direction only. An equal weight does not give a zero vector but a vector on the equator. Thus, they cannot be opposite quantities. Yet, the mathematical definition of orthogonality (inner product = 0) suggests an orthogonal relationship ($\langle 0|1\rangle = \langle 1|0\rangle = 0$).

Therefore, for generalization in quantum systems, one could interpret orthogonal quantities physically as those without any alikeness, not even opposition. Their coexistence would not affect one another, i.e. there would be no interaction. The advantage of having a complete set of orthonormal bases (orthogonal quantities of unit length) allows the *unique* description of *all* entities in the region. In the Cartesian coordinate system, the $x$, $y$ and $z$-axes (directions) are orthogonal to one another. For photons, polarization directions (left-/right-handed) are orthogonal quantities. In spintronics, it is not so much the charge of the electrons that carries information but its magnetic 'spin' property, comparable to spinning tops that turn left or right (angular momentum). In chemistry, a solvent is orthogonal to another when it does not dissolve the layer of material deposited from the other solvent. In our case, we work with superposition of individual information-carrying electric current running along different neural pathways. For example, a neuro-signal may run through both the limbic system and the cortex along (mathematically) orthogonal pathways. Their coexistence does not lead to interaction in a mathematical sense.

When additional constraints of probability of a quantum state with respect to other states are imposed, quantum entanglement is brought into the system. For instance, electron spins are simply orthogonal. It is the Pauli exclusion-principle that contributes to the quantum entanglement of two electrons at the same energy state.

Designed like this, a quantum processor has three logic gates (comparable to a conventional computer): The AND gate brings a bit into superposition, the OR gate changes the axis of, for instance, the spin of a qubit, and the Controlled-NOT or CNOT reverse-codes the input of unequal values: `1` to `0`, `0` to `1`. A controlled gate such as the latter works with two qubits. In the case of NOT, it operates on the second qubit when the first qubit is `1` (or in Dirac notation: $|1\rangle$, see the section named *Psychological State Vector*). Else, NOT leaves the second qubit untouched.

[4] In Eastern cultures, the Yin-Yang principle (☯) would illustrate the idea.

## A Bloch Sphere of Emotions

To model the ambiguity or 'polyvalence' of human emotions, such that a robot could simulate them in a human-like fashion, Raghuvanshi and Perkowski (2010) as well as Yan, et al. (2015) formalized emotion and emotional intensity by means of a Bloch sphere (Bloch, 1946). This ambition resembles that of Russel and Carroll (1999a, 1999b), plotting six clusters of affect items on a 2-dimensional circle with two axes of valence and activation. A circle representation also is offered by Scherer (2005), locating distinctive emotions in a dimensional structure of valence and arousal.

A Bloch sphere is a ball-shaped geometrical account of the pure state space of a quantum physical system that has two levels (i.e. a qubit). Put simply, a Bloch sphere represents a coordinate system of three axes (*x*, *y*, *z*) within a transparent ball (Figure 1) in which the 'arrow' or vector coming from its origin points out the 'state' that the system is in. That point $|\Psi\rangle$ (psi) may lie inside the ball or at its surface (Figure 1). For any point on the surface of the Bloch sphere, a state vector $|\Psi\rangle$ is handled by angle $\varphi$ (phi) for left-right and back-forth movement over the *x*- and the *y*-axis and by angle $\theta$ (theta) for up-down movement over *z*.

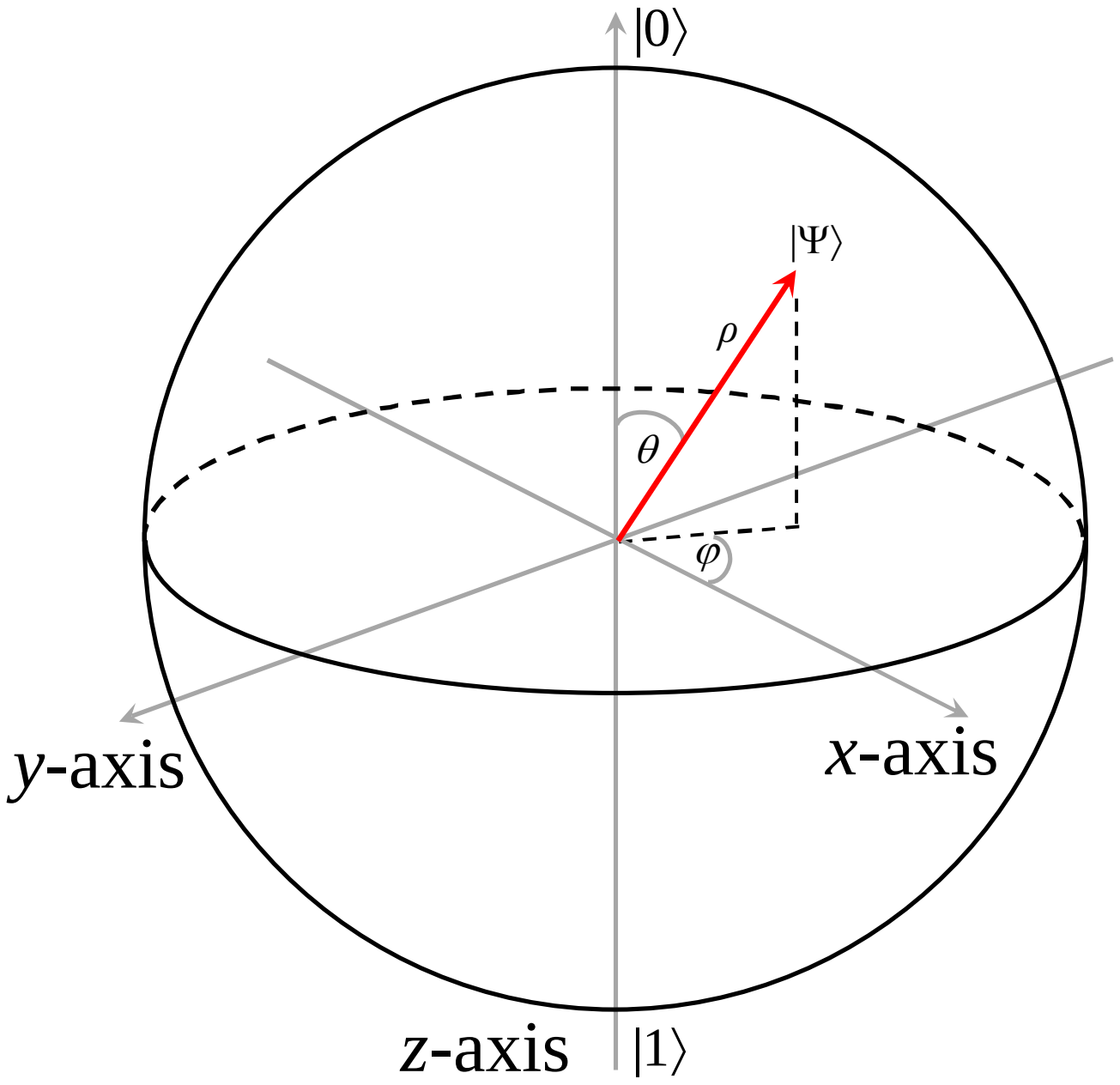

Figure 1: Bloch sphere representation of a qubit.

For instance, Raghuvanshi and Perkowski (2010) envisioned the active-passive dimension of human behavior along the *x*-axis and positive-negative emotions along the *y*-axis, the combination of which would point out a number of discrete emotions such as joy and anger. In their account, the *z*-axis represents the intensity of the emotion that is designated by the *x*,*y* coordinates. This "quantum sphere of emotions," as Raghuvanshi and Perkowski called it, allows for the occurrence of an ensemble of emotions rather than one single occurrence of one single emotion. Observation of just one emotion only would happen after measurement, for example, when a researcher runs a questionnaire or measures fMRI.

In the wake of Raghuvanshi and Perkowski (2010), Yan et al. (2015) also represented human emotion through geometry, namely as a qubit that defines a point on the Bloch sphere to express emotional ambiguity. These authors discerned on the *x*-axis a psychological dimension that goes from displeasure to pleasure. On the *y*-axis, they plotted a dimension going from sleep to arousal. The *z*-axis indicated the intensity of the emotion, going from more intense (1) (or in Dirac notation: $|1\rangle$) to less intense (0) or $|0\rangle$. With the intersection or 'cross point' of axes as the indicator of a 'neutral' state, emotions become stronger the more they move away from the origin. Represented like this, the emotion 'surprise' may have different levels of liveliness and may have pleasurable as well as unpleasable aspects concurrently.

In Yan's et al. (2015) rendition of expressing an emotion, a quantity of information (i.e. a qubit) pertains to the two angles $\varphi$ and $\theta$, in which $\varphi$ ($0 \leq \varphi \leq 2\pi$) represents an emotion such as happiness or sadness and $\theta$ ($0 \leq \theta \leq \pi$) specifies its intensity. The upper bound of angle $\varphi$ can be anywhere between pleasure and displeasure so that $\varphi$ may represent various (mixes of) emotions with different levels of arousal (i.e. away from the origin is stronger) as indicated by $\theta$.

## Affect and Reflection

Emotions are not the mere outcome of an affective process. Sometimes people wilfully alter the affective response, for instance, through mindful meditation or cognitive reflection (Schwartz, Stapp, & Beauregard, 2005). Empathy is said to be a 'cognitive emotion' because on the one hand a person has to take the perspective of someone else (a cognitive operation) and on the other hand imagine what that person might feel (an act of affect). Moreover, emotion-regulation strategies are nothing but gaining cognitive control over otherwise too intense affective responses.

Suppose something happens that upsets a person. For example, the owner of a robot observes how it trips over and breaks - as robots often do. The owner 'feels sorry' for the robot and attributes it 'real pain' (cf. Konijn, Walma van der Molen, & Van Nes, 2009). The information of the fallen robot enters the owner's Thalamus. On a neurological level, information is forwarded by firing frequencies of electrons or 'electric oscillations.' The Thalamus works as a semi-transparent mirror: Information runs directly to the Amygdala; psychologists would term this 'affective processing;' and *concurrently* that same information splits off to the Neocortex and only then enters the Amygdala, which psychologists would regard as 'reflective processing' (LeDoux 1996/1999; Cerić, 2012; Cone & Konijn, 2018). It follows that the information entering the Amygdala is present in more than one state all at the same time. In the case of the stumbling robot, information would directly enter the Amygdala to detect danger of falling and the detour through the Neocortex would assure that this danger does not concern a human being but a non-living thing: 'It's only a robot.'

What we should model is the observation that information is processed in parallel (cf. Crone & Konijn, 2018), by both affective and reflective brain circuits but that suddenly, the reflective system may be blocked or the affective system subdued (although not completely). Next, we deviate from Raghuvanshi and Perkowski (2010) and Yan et al. (2015) by the introduction of not just affective but also reflective processes and how these interact. The affective process we model does not stem from an arousal-pleasure-action theory so much but rather from the concern-driven theory by Frijda (2006), where emotional relevance (i.e.

importance, urgency) and valence (i.e. outcome expectancies) play a pivotal role. We also make a case that what we model is not mere mathematics but the actual physical basis of mental processes that happen in parallel in the brain.

## Psychological State Vector

Picking up on the example of our startled robot owner, there is a fast lane for electrons that go from Thalamus to Amygdala (Cerić, 2012). Let us call this oscillation over the $\hat{y}$-axis.[5] And at the same time, there is a slower-going stream running from Thalamus via the Neocortex to the Amygdala (Cerić, 2012). Let us call this oscillation over the $\hat{x}$-axis (Figure 2). Together, as depicted in Figure 2, they form a cortico-basal ganglia-thalamo-cortical loop (Yager, Garcia, Wunsch, and Ferguson, 2015), involved in the experience of reward as well as fear.

Based on this flow, we define a two-state system with the information processed affectively (the fast lane) described by eigenstate $|0\rangle$ and that processed reflectively (the slow lane) described by eigenstate $|1\rangle$ (Figure 2). In Dirac notation, these states are called *ket 0* and *ket 1* (Dirac, 1958). Another way of writing this is in matrix notation, where a state can be represented by a vector $\binom{\text{affective}}{\text{reflective}}$, implying that $|0\rangle = \binom{1}{0}$ and $|1\rangle = \binom{0}{1}$. *Eigenstates* are the quantum states observed when the quantum system collapses during measurement so that either the affective aspect *or* the reflective aspect is observed but not both, or something in between.

For every *ket*, there is a one-to-one corresponding dual vector denoted as $\langle n|$ ($n = 0, 1$), the Hermitian conjugate of $|n\rangle$, which Dirac (1958) would call a *bra* vector. A *bra* carries exactly the same information as a *ket* but is particularly useful for mathematical formulation.

Eigenstates are orthonormal to one another. By measurement, we see but one aspect while the other eludes us. Mathematically, the orthogonal property is defined by their zero inner product, i.e. $\langle 0|1\rangle = \langle 1|0\rangle = 0$, i.e. an eigenstate has no projection on another. As depicted in Figure 2, these eigenstates underlie the experience of reward and fear. The idea is that information moves into the direction of a certain interpretation (affect, reflection) via **self-observation** and/or observation by others, which could be thought of as an 'entanglement of perspectives.'

[5] To show we are working in a normed vector space with a spatial vector of length 1, we write a 'hat' on *x*, *y*, and *z*.

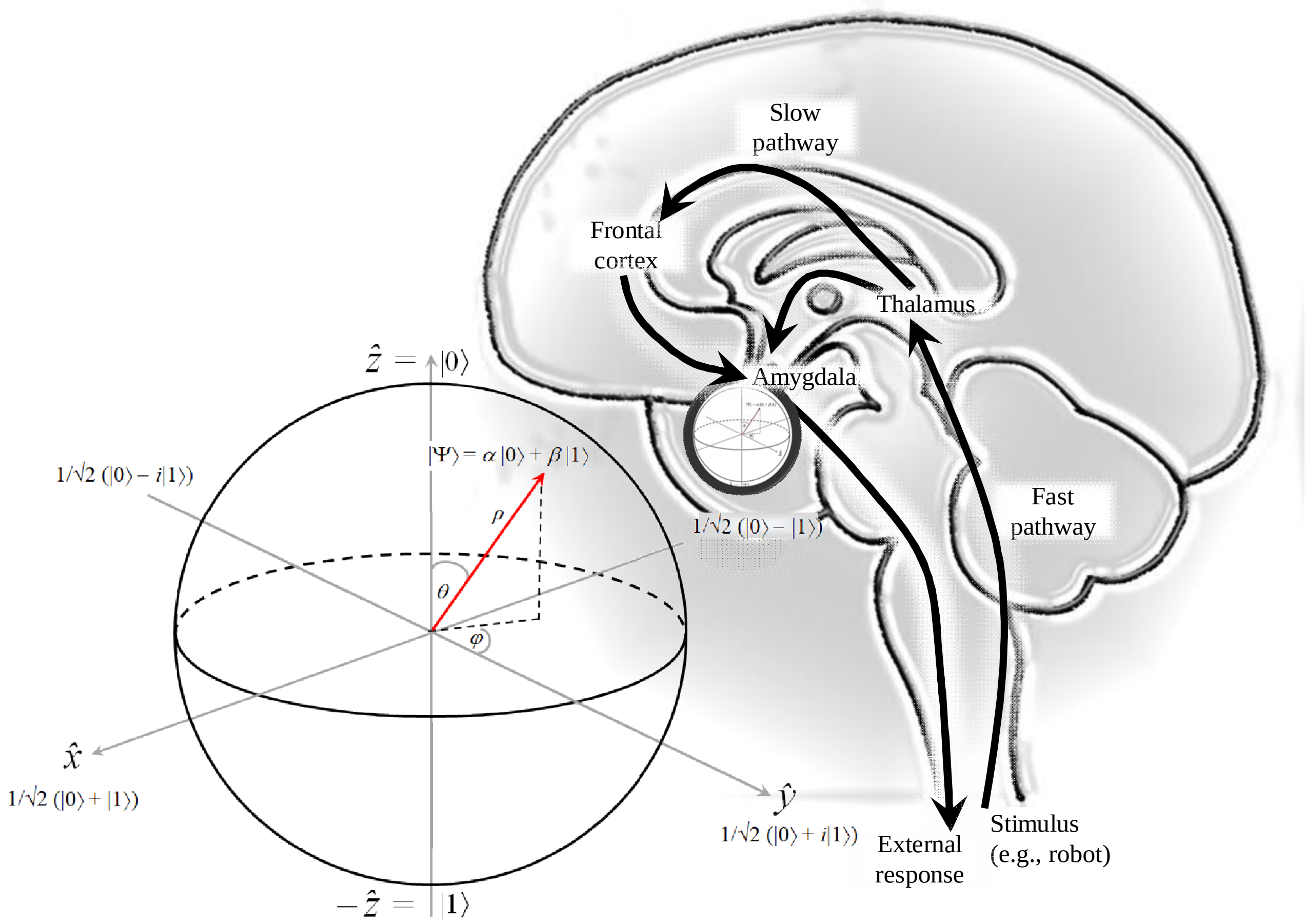

Figure 2: Amygdala represented by Bloch sphere with psychological state vector $|\Psi\rangle$.

When unobserved, however, the degree to which the Amygdala is simultaneously hit by electric oscillation via the fast affective and via the slow reflective route may differ. Owing to mental or environmental influences, the Thalamus may let more information pass through the affective route than through the reflective route or v.v. We model a *pure* psychological state $|\Psi\rangle$ (psi) within the Amygdala as a degree $\alpha$ where oscillations go through the affective route and a degree $\beta$ through the reflective route, stimulating *a certain location* in the Amygdala (see the section *Relevance, Valence | Reflection*), represented by (cf. Figure 2):

$$|\Psi\rangle = \alpha\,|0\rangle + \beta\,|1\rangle \qquad (1)$$

where $\alpha, \beta \in \mathbb{C}$. We use complex numbers ($\mathbb{C}$) because proportions of $\alpha$ and $\beta$ could differ by a phase which is not observed but estimated and thus is ‘imaginary’ (see section *Measurement: Imaginary, Real, and the Anger-and-Joystick*). The probabilities of becoming $|0\rangle$ and $|1\rangle$ should sum to unity, given the normalization condition $\langle\Psi|\Psi\rangle = |\alpha|^2 + |\beta|^2 = 1$. Conventionally, this is achieved by filling in $\cos(\theta/2)$ for $\alpha$ and $\sin(\theta/2)e^{i\varphi}$ for $\beta$ (e.g., Williams, 2011, pp. 11-13).

Now the state the Amygdala is in can be represented as a vector (red arrow in Figure 2) of the two eigenstates in a Bloch sphere (Bloch, 1946). Angle $\varphi$ (phi) is measured from the positive *x*-axis in a counter-clockwise manner and controls the ‘left-right’ and ‘back-forth’ movement of the state vector $|\Psi\rangle$ with values between $0 \leq \varphi \leq 360°$ ($2\pi$ rad).[6] Angle $\theta$ (theta) is measured from the positive $z$-axis and regulates the 'up-down' movement of the vector with values between $0 \leq \theta \leq 180°$ ($\pi$ rad). This renders a qubit of the geometry shown in (2), indicating some point on the Bloch *surface* (cf. Yan et al., 2015):

$$|\Psi\rangle = \cos(\theta/2)|0\rangle + \sin(\theta/2)e^{i\varphi}|1\rangle, \qquad (2)$$

or, in matrix notation:

$$|\Psi\rangle = \begin{pmatrix} \cos\frac{\theta}{2} \\ e^{i\varphi}\sin\frac{\theta}{2} \end{pmatrix}.$$

As said, we assign the reflective part of information processing to the $\hat{x}$-axis and the affective part to the $\hat{y}$-axis, projected on the Bloch sphere. Except when information is in the states $|0\rangle$ and $|1\rangle$, all the points (states) on the surface of the Bloch sphere pertain to superpositions of $\alpha|0\rangle + \beta|1\rangle$ with a unique set of $(\theta, \varphi)$. If $\theta = 0°$, $|\Psi\rangle = |0\rangle$; if $\theta = 180°$, $|\Psi\rangle = |1\rangle$. The $+\hat{z}$ and $-\hat{z}$ poles are $180°/2 = 90°$ apart in the Hilbert space (i.e. orthogonal as discussed), and the $\hat{z}$ and $\hat{y}$ states are $90°/2 = 45°$ apart. The surface of the Bloch sphere denotes all the 'pure states' that the information can be in (combinations of affective $|0\rangle$ and reflective $|1\rangle$).

Empirically, however, a more ambiguous situation occurs: Various psychological states are activated in tandem. These states form a statistical mixture called a 'mixed state' $|\Psi\rangle$, in which many pure states that the information or electric oscillations can be in are involved with some probability distribution. The Amygdala is hit by sets of activations, each having its own superposition of affective and reflective processing. We can express this multitude of oscillation pathways entering Amygdala, using the density matrix $\hat{\rho}$ (rho) (e.g., Fan, Peng, Zhang, Liu, Mu, & Fan, 2019):

$$\hat{\rho} = \Sigma_k \, p_k \, |\Psi_k\rangle \, \langle\Psi_k| \, , \qquad (3)$$

where the non-negative $p_k$ is associated with the probability of getting the pure state $|\Psi_k\rangle$ in the ensemble of the system such that $\Sigma_k \, p_k = 1$.

To determine from its explicit form whether a quantum state of the signal transportation in the Amygdala is pure or mixed, the sum of eigenvalues of $\hat{\rho}$, i.e. $\mathrm{Tr}(\hat{\rho})$, is equal to 1 for pure states and smaller than 1 for mixed states. In terms of Bloch sphere representation, mixed states are located *inside* the Bloch sphere and different points can be expressed with a set of $(\rho, \theta, \varphi)$ in $\mathbb{R}^3$ (the real coordinate space in spherical coordinates), where $\rho$ ($0 \leq \rho \leq 1$) (rho) is the length (magnitude) of the state vector $|\Psi\rangle$. Points of $|\rho| = 1$ indicate pure states and $|\rho| < 1$ indicate mixed states. The origin $\rho = 0$ implies the so-called completely depolarized state.

That means the following: In Figure 3 (left panel), an information state possesses a state vector $|\Psi\rangle$. Based on this constellation, the vector $|\Psi\rangle$ now specifies a coordinate in $\mathbb{R}^3$, indicating a point of the Bloch sphere that shows the psychological state Amygdala is in. The information travels concurrently through the affective route with the state $|0\rangle$ and the reflective route with the state $|1\rangle$, under the set of bases from the measurement of $\hat{S}_z$. These two bases are represented by the positive and the negative *z*-axes respectively in the Bloch sphere, which are fixed in direction. Such measurement acts as a projection of the state vector on the *z*-axes. On the other hand, the *same* |Ψ⟩ can also be expressed on the basis of reflection and affect (analogous to the *xy*-plane), and the corresponding operators of measurement are $\hat{L}_{xy}$ and $\hat{L}_z$. The direction of the projection on the *xy*-plane are also fixed. If the psychological state now changes (transforms) to |Ψ'⟩, and measurements of the state are carried out again, the magnitude along the *z*-axis (affective/reflective route) as well as that on the *xy*-plane (state of reflection and affect) also changes, leading to a different apparent scaling of the projection as shown in Figure 3 (right panel). Hence, the state vector |Ψ⟩ at time 1 indicates a different interior point in the Bloch sphere (in $\mathbb{R}^3$) than it does at time 2 (state vector |Ψ'⟩). Note that the vector bases (bases on the *xy*-plane and along the *z*-direction) are not affected by the scaling; they only represent the direction of the projection (the measurement to be taken) and so stay under the same orientation as before. However, |Ψ⟩ *changes* through the transformation and becomes vector |Ψ'⟩.

[6] We follow the physics and engineering conventions here, not mathematics.

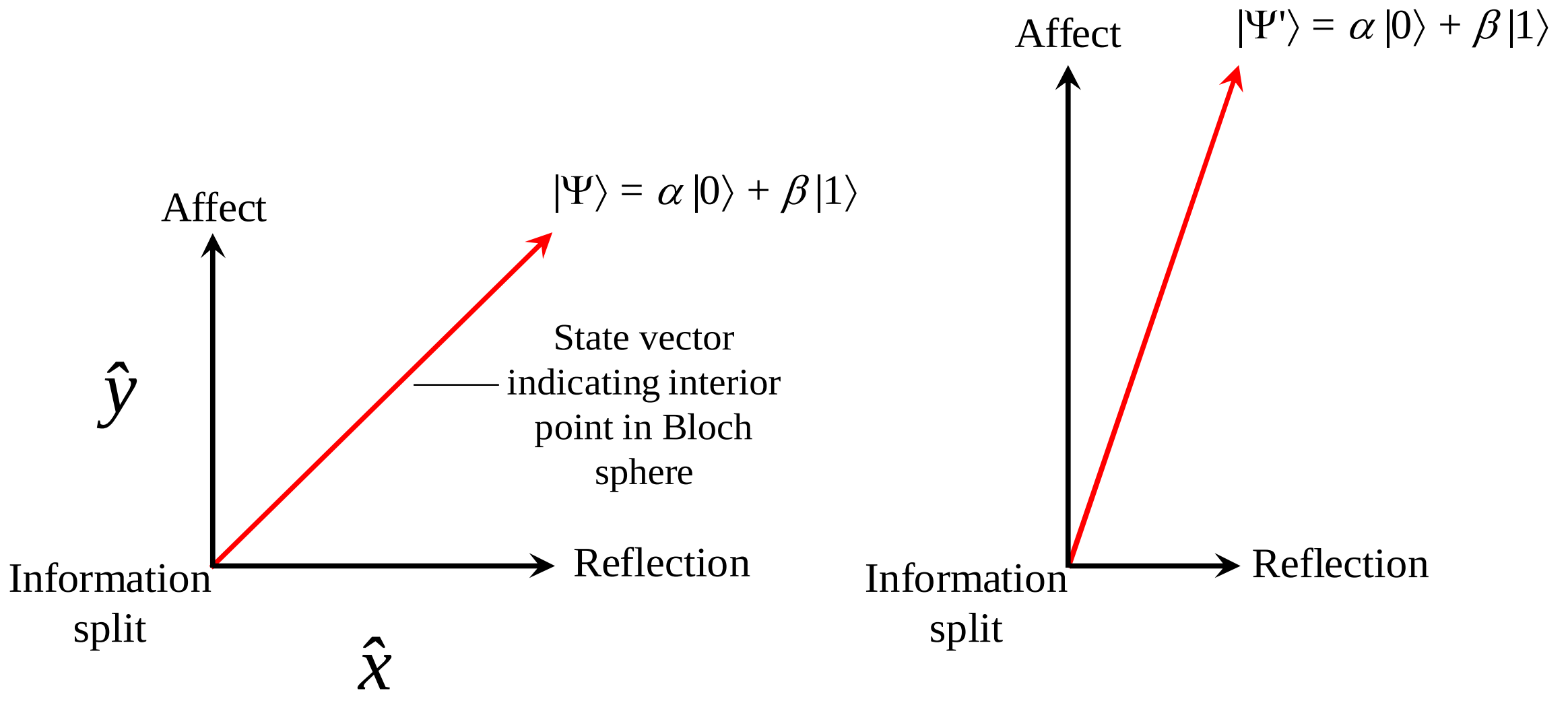

Figure 3: Directions of vector x̂ and vector ŷ (their angle) are not affected by scaling or other transformations but direction of vector |Ψ⟩ is (after Spruyt, 2014).

If vector |Ψ'⟩ at time 2 happens to be a multiplication of vector |Ψ⟩ at time 1 (i.e. the way information is processed is a multiplication of itself), then vector |Ψ⟩ is an eigenvector of the system. And the *number* by which eigenvector |Ψ⟩ is multiplied is the related eigenvalue, which may be 0 but can also be 1. If eigenvalue is 1, this means that psychological state vector |Ψ⟩ is stable and does not change its direction over time no matter how often it is multiplied. The sum of eigenvalues of $\hat{\rho}$ should equal 1 in pure cases and be smaller than 1 in mixed cases, indicating the pure (affective *or* reflective) or mixed (affective *and* reflective) state the information in the Amygdala is in.

## Relevance, Valence | Reflection

In emotion psychology (e.g., Frijda, 2006), emotions are the result of two affective dimensions. When an event occurs, *Relevance* indicates the gravity, severity, urgency, or importance of that event to the goals and concerns of the agency. Something of great importance exerts highly intense emotions, whether in joy or fear. *Valence* is an affective dimension that indicates the direction of the emotion. When something facilitates goals and concerns, emotions are positive; when inhibiting those goals and concerns, they will be negative. Related action tendencies are positive approach (e.g., to hug) for positive valence, and fight or flight for negative valence, while sitting still and do nothing (‘freeze’) may come from both.

**Relevance: bursts of electrons**

The particle that carries the information through the brain structures is the electron. Between axon and dendrite, neurotransmitters are released from synapse to receptor cell to facilitate the electron transfer process (e.g., Taherpour, Rizehbandi, Jahanian, Naghibi, & Mahdizadeh, 2015). The frequencies of releasing electrons or ‘firing frequencies’ carry the information and the more electrons run at a location with a certain functional specialization, the more energy the psychological system residing there consumes (e.g., language, memory). In other words, the psychological experience of gravity, severity, urgency, or importance of an event may be the amplitude or magnitude of the signal, indicating the number of electrons rushing through that location, i.e. the density of the electron burst, and the height of the energy level. Relevance, then, would be the electromotive force (consistent with “density”) by which information is pushed through, in our case, the Limbic system and (less so) the Neocortex. When the magnitude is relatively high, according to an internal ‘potentiometer,’ the event is ‘relevant.’ When the magnitude remains low, the event is deemed ‘irrelevant.’ In equation (1), the $\alpha$ and $\beta$ of psychological state vector $|\Psi\rangle$ represent the Relevance of an event to a person’s goals and concerns, physiologically measurable as an electromotive force.

**Valence: position on the $\hat{y}$-axis**

The parallel occurrence of positive and negative affect has always been problematic to explain in psychology, which has come to expression in surveys that run bipolar rating scales and semantic differentials for measurement (either positive or negative). In cases where researchers ran multiple unipolar scales to measure opposites, it showed that affective states are not necessarily bipolar (positive = 1 – negative). However, the parallel occurrence of ambiguous states was inferred (i.e. an *ex-consequentia* fallacy) rather than formally accounted for (i.e. *modus ponens*).

Although Amygdala is a small brain structure, it is composed of several units and pathways with their own function. For example, rewards and positive valence are associated with the lateral nucleus of the Amygdala whereas aversion, fear, and negative valence are related to its central nucleus (Wilensky, Schafe, Kristensen, & LeDoux, 2006). It is also found that beyond Amygdala, the posterior left hemisphere seems to be functional in collaboration with the orbitofrontal cortex and ventral striatum for processing negatively valenced stimuli (i.e. facial expressions, see Killgore & Yurgelun-Todd, 2007). When projected in the right visual field, negatively valenced stimuli seemed to activate left anterior regions, whereas positively valenced stimuli activated the right anterior regions. Additionally, these authors found an interrelated system in the posterior right hemisphere for emotional perception in general but also with a special sensitivity for negatively valenced facial expressions.

We forward that specialized areas in the brain interact with each other and that information (i.e. electrons) can be in superposition in all those areas important to, for example, processing emotions. An ambiguous stimulus (e.g., food that hurts while eating) activates the lateral *and* the central nucleus but may also reach the Neocortex (i.e. the frontal lobe), overthinking the mixed emotions.

In other words, oscillations over the $\hat{y}$-axis are literally about the spatial position of electrons, for instance, being at the lateral side and inducing a sense of reward and/or being at the central nucleus and inducing a sense of fear. In Figure 2, then, the $\hat{y}$-axis represents Valence with its positive side on the Bloch sphere surface leading the information state $|\Psi\rangle$ to $|+\hat{y}\rangle = 1/\sqrt{2}\ (|0\rangle + i|1\rangle)$ and its negative side $|-\hat{y}\rangle = 1/\sqrt{2}\ (|0\rangle - i|1\rangle)$.

**Reflection: position on the $\hat{x}$-axis**

The Neocortex is known to harbor executive functions and control procedures. These may be more practical and 'shallow' such as planning and error correction or 'deeper,' such as philosophical thought. The simpler type of reflection may be the trained routine responses or quick-fix solutions to known situations, for example, when the teacher tells his or her student to solve a problem differently (cf. 'feedback'). The opposite would be an elaborate contemplation of possibilities and perspectives (cf. 'mindfulness'), being less focused on a local solution but rather on a global consideration of alternatives. Thus, on the side of reflection, more global and meditative forms of information processing may gradually change into more local, automatic, and conditioned thoughts about how to deal with a situation and v.v.

During the operation of epilepsy patients, Haller (2016) monitored the electrical activity of cortical neurons with electrocorticography (ECoG), thus increasing the spatial resolution of the measurement as compared to conventional EEG. Haller found that the prefrontal cortex mainly had a coordinating task over other brain areas (e.g., memory, language) while formulating a thought. With a difficult question that has no routine answer, the brain has to think harder, which is visible in increased brain activity (fMRI). However, Haller (2016) found that this was not due to an increased firing frequency of the respective neurons but to the recruitment of more areas of the cortex to crack the problem. In other words, increased brain activity in the cortex indicates more multitasking.

Thus, Reflection may have two ends, the more shallow 'routine response' to known situations or the more thoughtful elaboration or 'contemplation,' which draws on more sources of information than routine answers do. The difference is the number of areas that are involved into producing the reflection.

Oscillations over the $\hat{x}$-axis, then, literally indicate the spatial position of electrons, for instance, addressing memory locations alone (routine response) or additionally involving functional regions of the prefrontal cortex such as orbitofrontal higher-order reasoning and suppression of action (contemplation). In Figure 2, let the positive $\hat{x}$-axis represent Reflection with its multitasking side (contemplation, deep thought) leading $|\Psi\rangle$ to $|+\hat{x}\rangle = 1/\sqrt{2}\ (|0\rangle + |1\rangle)$ and negative $\hat{x}$-axis its 'mono'tasking side (routine response, shallow thought) leading to $|-\hat{x}\rangle = 1/\sqrt{2}\ (|0\rangle - |1\rangle)$.

In sum, let the states at $+\hat{x}$ and $-\hat{x}$ in Figure 2 stand for Reflection (i.e. deep thought) and routine responses (e.g., shallow thought), respectively. Let the $+\hat{y}$ and $-\hat{y}$ states stand for positive valence (e.g., leading to joy) and negative valence (e.g., leading to fear), respectively. Within this framework of opposite physical forces that the Bloch sphere describes, the information-carrying electrons may oscillate more into the direction of $+\hat{z} = |0\rangle$, the fast affective lane, or into the $-\hat{z} = |1\rangle$ direction, the slow reflective pathway. Yet, they do not do so completely; the electrons are distributed like clouds or clusters along that direction. When measurement through EEG, EcoG, or fMRI finds that activation mostly moves into the $-\hat{z}$ direction, it may also partially move into the $+\hat{x}$ *and* $-\hat{x}$ direction. With respect to Figure 2, then, the information that enters the Amygdala is described by:

| **State information** | **Psych. dimension** | **Observable effect** |
|---|---|---|
| $\lvert +\hat{x}\rangle = 1/\sqrt{2}\,(\lvert 0\rangle + \lvert 1\rangle)$ | Deep reflection | Thinking something over |
| $\lvert -\hat{x}\rangle = 1/\sqrt{2}\,(\lvert 0\rangle - \lvert 1\rangle)$ | Shallow reflection | Routine response |
| $\lvert +\hat{y}\rangle = 1/\sqrt{2}\,(\lvert 0\rangle + i\lvert 1\rangle)$ | Positive valence | Direction of affect is upbeat |
| $\lvert -\hat{y}\rangle = 1/\sqrt{2}\,(\lvert 0\rangle - i\lvert 1\rangle)$ | Negative valence | Direction of affect is down |
| $\lvert +\hat{z}\rangle = \lvert 0\rangle$ | Affective processing | Fast |
| $\lvert -\hat{z}\rangle = \lvert 1\rangle$ | Reflective processing | Slow |
| $\alpha\,\lvert 0\rangle$ | Relevance of affect | Tendency of emotion |
| $\beta\,\lvert 1\rangle$ | Relevance of reflection | Tendency of thinking |

Note that the observable effect comes with measurement. Higher relevance exerts a stronger tendency to execute desired response. Relevance is high if a goal or concern is worth the trouble of investing, putting effort into it. This relevance also comes with an intensity, which leads to the intensity of the emotion or thinking. Regarding the 'intensity of thinking,' we mean the energy or strength it takes to think. It would correspond to the density of electrons passing through the activated areas. Seen as energy depletion or effort, the intensity of thinking is the electric 'tension' or 'pressure' that happens at the process, measurable in μV as the difference in electric potential between two points on a (bundle of) cortical neuron/s. Depth of thought or deep reflection relates to the higher number of aggregated neuronal resources as compared to shallow thought (cf. the multitasking cortex, Haller, 2016). Depth and intensity may be correlated but not necessarily dependent on one another.

## Quantum Transition Matrices and Logic Gates to Express Change

So far, we discussed mixed psychological states that did not become entangled yet but we did not model how variations in reflection and affect happen continuously (cf. mood swings, change of mind). We will adhere to Yan, et al. (2015) in their use of emotion-transition matrices to account for psychological shifts. At one point, however, a 'choice' in the constellation is made and (in our case) affect and reflection become entangled. To describe how to transfer from one entangled state to another, we follow Raghuvanshi and Perkowski (2010), using quantum logic gates such as Einstein, Podolsky, and Rosen (EPR) circuits.

Raghuvanshi and Perkowski (2010) introduced the notion of quantum circuits that correspond to affective behavior. Figure 4 is a derivation from Raghuvanshi and Perkowski's proposal to employ EPR circuits to set qubits into an entangled state (Einstein, Podolsky, & Rosen, 1935). Figure 4 draws on a fuzzy-logics formalization of the affective system as simulated in a robot by Hoorn, Baier, Van Maanen, and Wester (submitted), relating to different appraisal dimensions such as identifying good and bad traits in another agency and building up a level of involvement with that agency in contrast to a level of emotional distance (Konijn & Hoorn, 2005; Hoorn, 2015). This is less of a long stretch than it seems. If appraisals and responses are regarded as 'a diagram of selective influences,' then Dzhafarov and Kujala (2012) demonstrate that two levels of the same psychological dimension can be treated as outcomes of non-commuting measurements performed on entangled particles as happens in EPR gates.

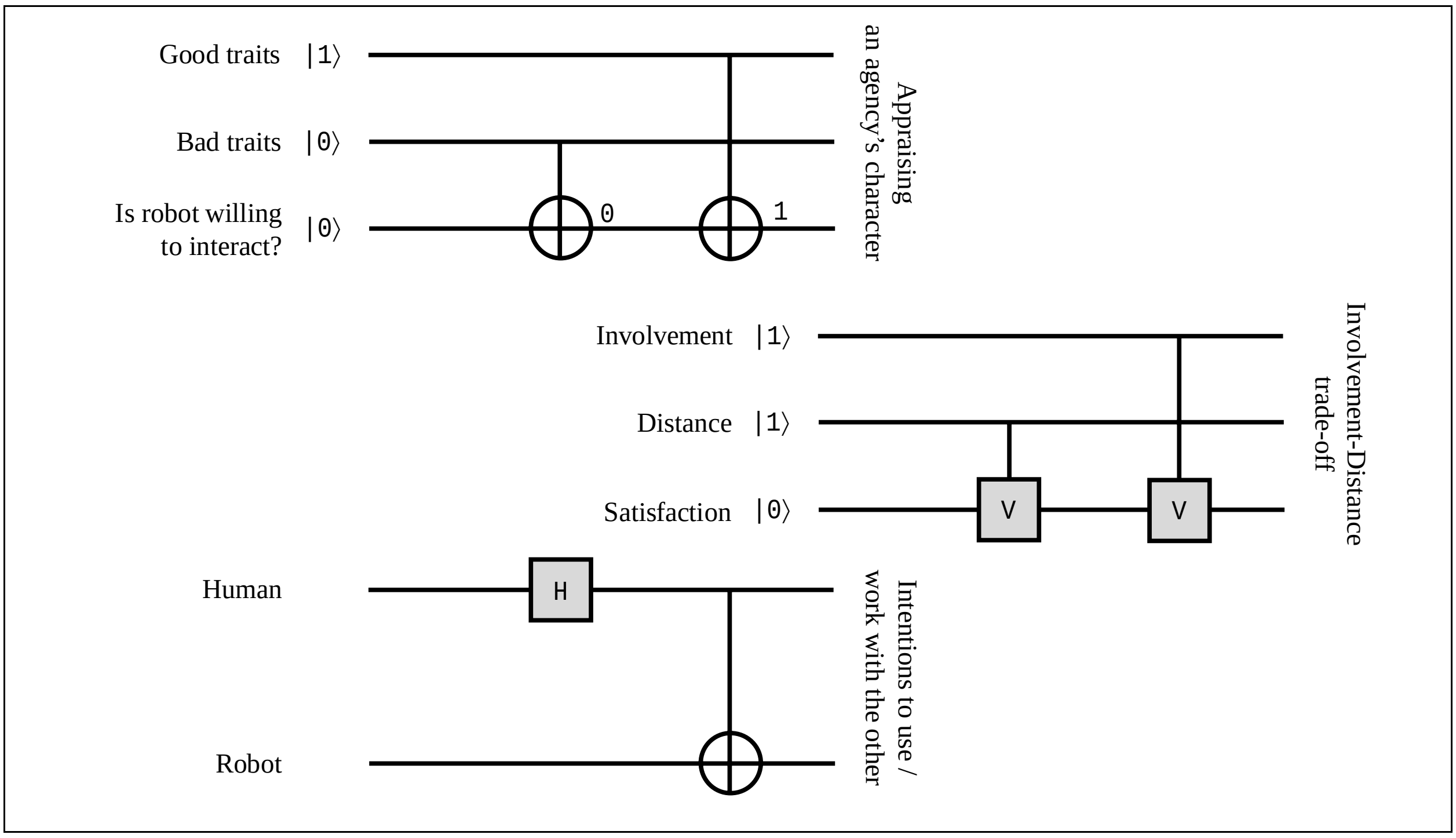

Figure 4: Quantum-logic circuits for a good-bad character, involvement-distance trade-off, and human-robot interaction (after Raghuvanshi & Perkowski, 2010).

In Figure 4, the affective behavior of the robot is designed such that the robot appreciates another agency (human or otherwise) when that agency basically has a good character (top line) – even when some other traits of that agency may be bad (middle line). Yet, when the good traits become entangled with the bad (e.g., steal to give to the poor), an ambiguous situation occurs and in this case, the robot decides it does not appreciate the other for stealing (bottom line). Table 1 provides the truth table for this circuit, supplemented with possible action tendencies.

Table 1: Truth table of an EPR circuit for an agency with good and bad traits.

| **Good traits** | **Bad traits** | **Interaction (initial: \|0⟩)** | **Action tendency** |
|---|---|---|---|
| \|0⟩ | \|0⟩ | \|0⟩ | Do nothing, sit still |
| \|0⟩ | \|1⟩ | \|1⟩ | Negative approach |
| \|1⟩ | \|0⟩ | \|1⟩ | Positive approach |
| \|1⟩ | \|1⟩ | \|0⟩ | Avoid |

The second situation in Figure 4 shows the use of a Controlled-V gate (Controlled-Square-Root-of-NOT) to model the parallel occurrence of a robot's involvement with its user as it is traded for affective distance (and v.v.), which determines the level of a robot's satisfaction with its user (Hoorn, Baier, Van Maanen, & Wester, submitted; Hoorn, 2015). Theory has it that either being involved or at a distance is not as satisfactory as mixes of involvement *and* distance (Konijn & Hoorn, 2005; Hoorn, 2015). Think of a surgeon who cannot keep professional distance because he is too involved with his patient (e.g., his own child) and does not want to do an incision. Table 2 provides the truth table for this circuit.

Table 2: Truth table of an EPR circuit for the involvement-distance satisfaction of an agency.

| Involvement | Distance | Satisfaction (initial: $\|0\rangle$) | Remarks |
|---|---|---|---|
| $\|0\rangle$ | $\|0\rangle$ | $\|0\rangle$ | Unsatisfied |
| $\|0\rangle$ | $\|1\rangle$ | $\frac{1}{2}\left[(1+i)\|0\rangle+(1-i)\|1\rangle\right]$ | In doubt |
| $\|1\rangle$ | $\|0\rangle$ | $\frac{1}{2}\left[(1+i)\|0\rangle+(1-i)\|1\rangle\right]$ | In doubt |
| $\|1\rangle$ | $\|1\rangle$ | $\|1\rangle$ | Satisfied |

Satisfaction depends on an Involvement-Distance trade-off. In the wake of Raghuvanshi and Perkowski's (2010) modeling approach, then, Figure 4 shows that it takes the parallel occurrence of Involvement *and* Distance to change the robot's Satisfaction with its user ($V\cdot V$ = NOT). If either Involvement or Distance is active then the robot is partially satisfied and partially not, i.e. the satisfaction qubit for the involvement-distance qubit registers $|10\rangle$ or $|01\rangle$ is $|S\rangle = \frac{1}{2}\left[(1+i)|0\rangle+(1-i)|1\rangle\right]$ with equal probability of having satisfaction of $|n\rangle$ ($n$ = 0, 1) of ($|\langle n|S\rangle|^2$ =) 0.5 (cf. the membership functions to a fuzzy set). When in Figure 4, Satisfaction is measured by a psychometric scale, the researcher observes 50% probability that the robot is satisfied and 50% probability that it is unsatisfied (i.e. mid-scale values). Only repeated measurement in independent copies of the Controlled-V circuit will reveal with higher probability whether the robot is satisfied, unsatisfied, or in doubt. Single measurement will not convey the states that Involvement, Distance, or Satisfaction are in. If the robot doubts whether it is satisfied or not, this means that one dimension in the Involvement-Distance trade-off is active. In the situation depicted by Figure 4, however, the state of Distance cannot be detected.

The third situation in Figure 4 employs an Einstein-Podolsky-Rosen gate to formalize an aspect of human-robot interaction, where both agencies develop Use Intentions based on measures of Valence (positive-negative outcome expectancies) (Van Vugt, Hoorn, & Konijn, 2009; Hoorn, 2015). The valences of the human and the robot before and after the interaction can be expressed as a qubit register |Human Robot⟩. In line with Raghuvanshi and Perkowski (2010), the following analysis would apply to the Human and Robot interacting:

$|0\rangle_H |0\rangle_R \rightarrow 1/\sqrt{2}\ |00\rangle + 1/\sqrt{2}\ |11\rangle$
$|1\rangle_H |1\rangle_R \rightarrow 1/\sqrt{2}\ |01\rangle - 1/\sqrt{2}\ |10\rangle$
$|1\rangle_H |0\rangle_R \rightarrow 1/\sqrt{2}\ |00\rangle - 1/\sqrt{2}\ |11\rangle$
$|0\rangle_H |1\rangle_R \rightarrow 1/\sqrt{2}\ |01\rangle + 1/\sqrt{2}\ |10\rangle$

If we define 1 as indicating positive Valence (e.g., "I expect that the robot is a great help") and 0 means negative Valence (e.g., "My user will not understand what to do"), probabilistic behaviors may be observed. Suppose a manager forces an employee to work with a robot but both human and robot have negative expectations about one another. According to the third circuit in Figure 4, two outcomes may result: There is a 50% chance that Human and Robot cooperate and have little expectations about it. There is also a 50% chance that they cooperate while both realize that their expectations are low and so they change this around into positive Valence (‘Let’s make the best out of a bad situation’).

According to theory (Van Vugt, Hoorn, & Konijn, 2009; Hoorn, 2015), variables such as Ethics (good-bad), Engagement (Involvement-Distance), and Use Intentions (willingness to use or not) are related and may occur in parallel. For example, an agency may be a bad person (e.g., a criminal), yet someone may feel empathetic towards him (an aspect of Involvement), and additionally may think that this person is useful for manual labor (i.e. Use Intentions). Thus, the states and transitions described above should be mapped onto a structure that describes the progression between states in the system state space.

State-transition networks consist of a graph of states that can occur simultaneously, within systems as well as between systems. States are connected by labeled arcs that indicate actions or transitions. State-transition networks have a start and an end state and are fit to describe, in our case, one-to-one human-robot interactions. Typically, transitions between states take place when facilitating or inhibiting operators are present (Luyten, Clerckx, Coninx, & Vanderdonckt, 2003).

In Figure 5, a state-transition network is shown for the activation of one state alone (upper left panel), a combination of two states operating concurrently (upper middle), or the occurrence of three states all together (upper right).

Transitions between states go back and forth through the respective quantum logic gates (Figure 5, lower panel), so that the robot may progress from standing idle to processing Ethical information (good-bad), Engagement (Involvement-Distance), and Use Intentions (willing to use or not) all at once; or the other way around, falling back into idle mode again.

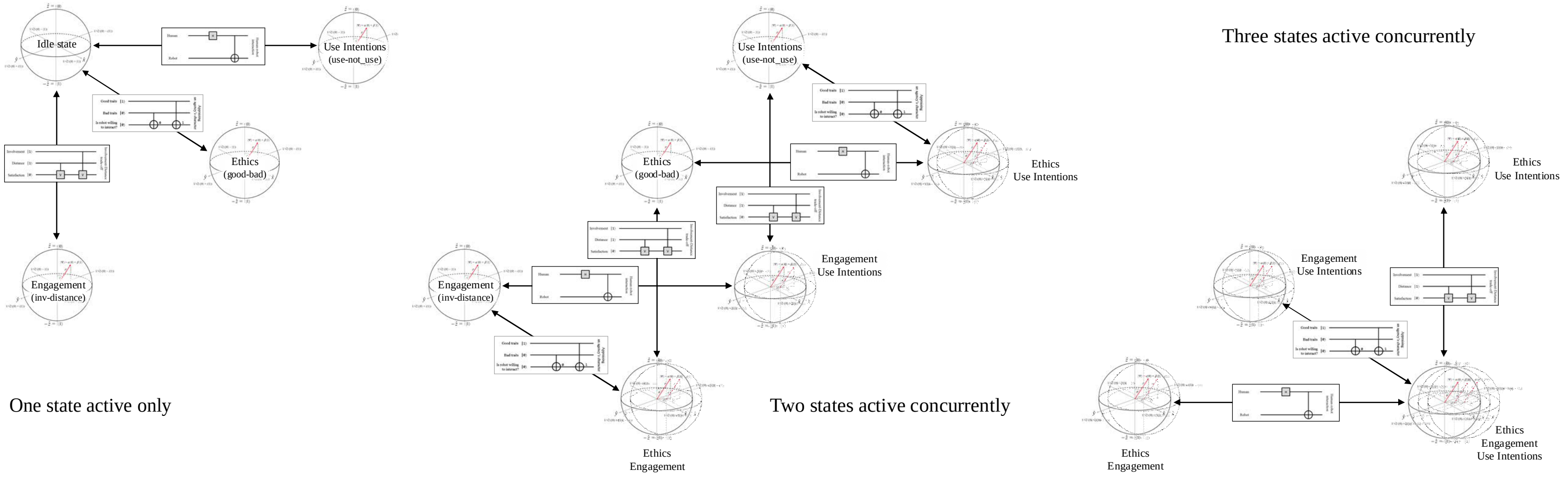

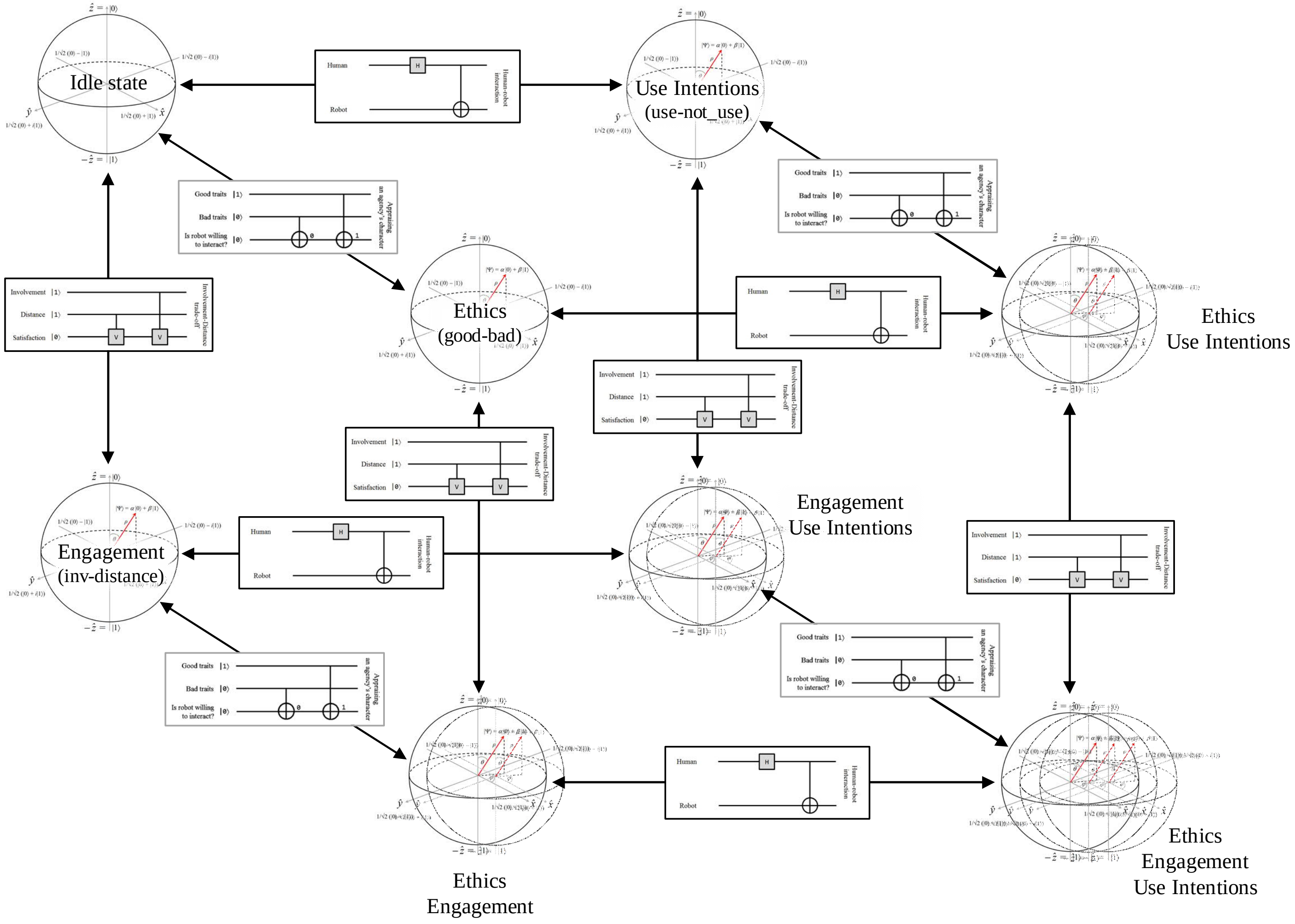

Figure 5: Quantum-state transition-network using logic gates for the appraisal of Ethics, Engagement, and Use Intentions (cf. Hoorn's, 2008, fuzzy hypercube).

Psychological states are concurrent rather than sequential. Therefore, in the design of a qubit account of a robot's social behavior, state transition-networks are most useful also to find dangerous states in its affective-reflective system.

The upper part of Figure 6, for instance, shows two states between which the robot can switch back and forth: Seeking affective interaction $|0\rangle$ and doing reflective interventions $|1\rangle$. Suppose the robot is dissatisfied with its user because it is maltreated (cf. Konijn & Hoorn, 2018) but may do its user no harm. Thus, the robot regulates its emotions. It may have a number of regulative actions from which it can make a choice (Hoorn, 2018a), one of which is to avoid the situation. If it chooses to avoid the situation, stop interaction, walk out on its user, and exit the scene, this intervention establishes a higher level of satisfaction than before. The dangerous state is in doing nothing (to 'freeze'). If no choice of action is made, no change in states occurs but this also keeps dissatisfaction at its prior (too high a) level and potential mitigation is not happening. Potential new points of mixed states that the psychological state vector $|\Psi\rangle$ might indicate are not 'kept' or secured. By doing nothing, possible changes in affect are not 'saved' as it were. No choice is made, everything remains as is.

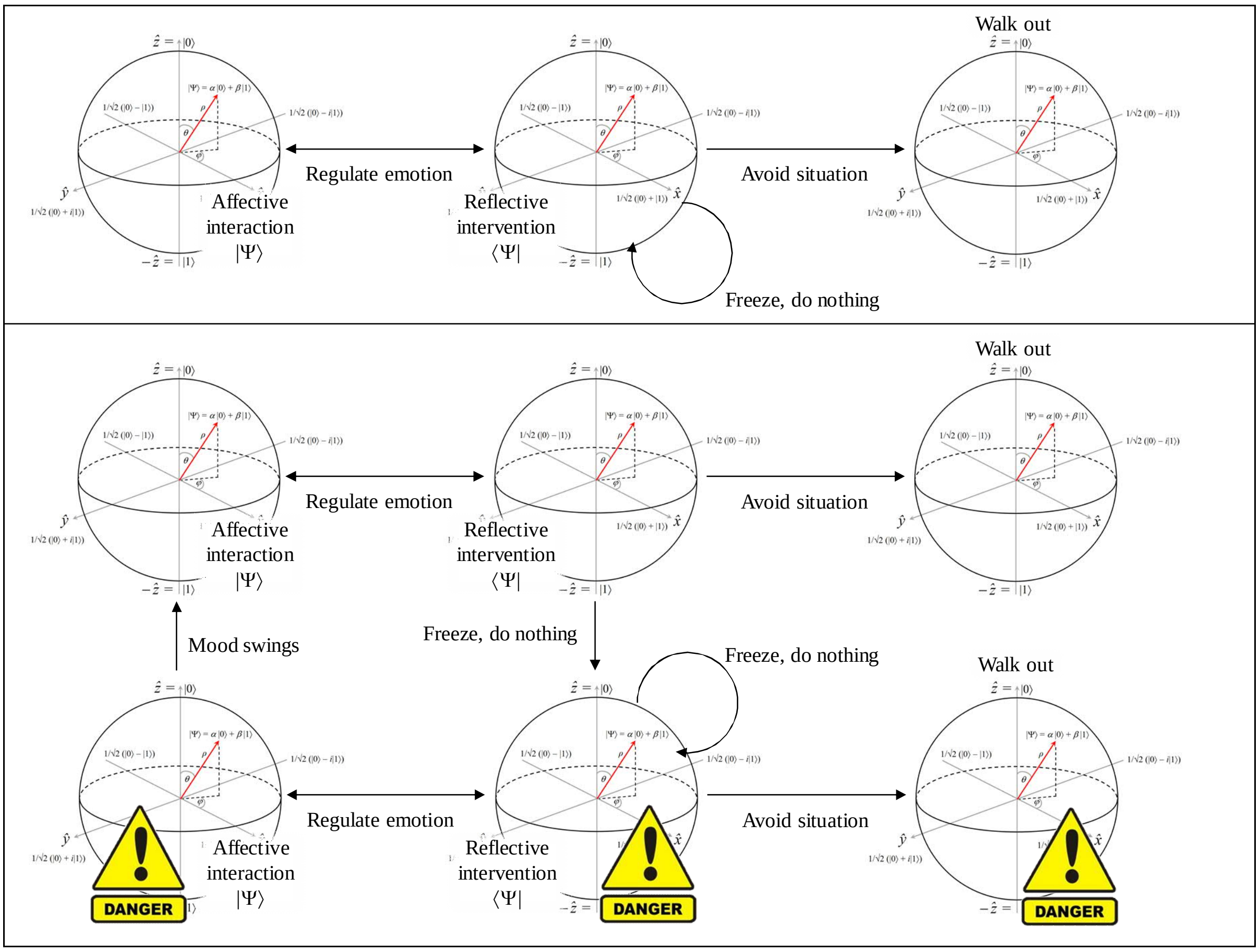

Figure 6: State-transition networks show potential dangerous states in the design of a robot's affective-reflective system.

The lower part of Figure 6 shows the design of the two states of affective interaction $|0\rangle$ and reflective intervention $|1\rangle$ as a duplicate system. In this set-up, it is possible for the robot to avoid interaction and walk out with or without adapting its original level of (dis)satisfaction. The without-part is the dangerous part, when the robot chooses to freeze and do nothing about its situation. At that point, no change in states happens. In this architecture, negative affect may remain unchecked and, for example, mood swings may aggravate the prior situation. However, by undertaking no action, potential new positions inside the Bloch sphere are lost. In that case, the interaction may be exited without satisfaction being positively adapted; it may even be so that while walking out the robot is worse off (due to the mood swings).

## Measurement: Imaginary, Real, and the Anger-and-Joystick

A major challenge in all of science, not just quantum physics, is that measurement interferes with the phenomenon under study. Textbook example is that position and momentum of a particle cannot be measured together; they are mutually exclusive or 'non-commuting.' An answer could be that as energy is mass in motion (i.e. 'momentum') and a measurement device tries to fixate its location in a snapshot, the attempt is inherently bound to fail since the energy moved on. The snapshot merely shows location with momentum having escaped already. Mathematicians now say about this particle that 'its wave function collapsed to eigenvalue,' a single measured value. However, it is not possible to observe that happening; one can merely calculate it, which is a guess or reconstruction at best.

The same is true for psychological methods. As soon as someone realizes s/he is observed, the behavior changes, and what one measures is psychological behavior while being observed in that particular (experimental) situation. Someone's psychology cannot be measured without changing the state a person is in. A questionnaire conducted during or after emotion induction forces the psychological state of the participant into 'reflection,' not 'affect.' Being made from metal, invasive brain probes but also non-invasive electrodes interfere with the electro-magnetism of the electric information currents.

Because observers and instruments are physical entities that together with their objects of study tie in as a complex of wave functions, it is impossible to deduct exact outcomes from measurements. That means we deal with mere probability distributions and so subjective decisions of the observers are part of the causal explanations of physical relationships (cf. Schwartz, Stapp, & Beauregard, 2005).

The fundamental observation problem has serious consequences for what we can say about the 'real' world. Because we have to 'guesstimate' what happens, there is an imaginary side to any scientific statement about the physical universe. In quantum approaches, this is solved by assuming a coordinate space of real numbers (Re in $\mathbb{R}^3$) but also of complex or imaginary numbers (Im in $\mathbb{C}^3$).

Figure 7 consists of three parts. The top segment shows a Bloch sphere supplemented with equations for the 'real' and imaginative part of measurement. The middle segment shows a framework to understand the epistemic consequences of a Bloch sphere representation that cannot be fully measured quantitatively. The bottom part offers the assimilation of the partially measured Bloch sphere by the observer's mental representation of the physical world.

The Bloch ball in the top section of Figure 7 indicates that the state measured on the $\hat{z}$-axis is in the observer's mental representation of the world a real observation (Re = real) of $\hat{x}$ folded together with an estimate or imaginary (Im and $i$) measurement of oscillations over the $\hat{y}$-axis. Obviously, the reverse observation of $\hat{y}$ as Re and $\hat{x}$ as Im may also be the case.

With Reflection on the $\hat{x}$-axis being a 'real' observation through an emotion-assessment questionnaire, the measurement scale for deep reflection on a person's emotions (1/√2 (|0⟩ + |1⟩) and the measurement scale for shallow thought about one's feelings (1/√2 (|0⟩ – |1⟩) have no 'imaginary' part, according to the observer. However, what actually happened emotionally deep inside that person eludes the observer and perhaps even the person self. Thus, oscillations over the $\hat{y}$-axis of the Affective system do have an imaginary component. In other words, the estimate of positive valence becomes $|+\hat{y}\rangle = 1/\sqrt{2}\ (|0\rangle + i|1\rangle$ and for negative valence $|-\hat{y}\rangle = 1/\sqrt{2}\ (|0\rangle - i|1\rangle$.

A measurement in quantum mechanics mathematically is formulated with an operator $\widehat{M}$. Thus, the emotion-assessment questionnaire can be regarded as an operator $\widehat{M}$ operating on the state ket $|\Psi\rangle$, donated as $\widehat{M}|\Psi\rangle$. The questionnaire is usually designed to obtain a number of possible values of a parameter such as valence. Then $\widehat{M}|\Psi\rangle = \sum_k \langle k|\Psi\rangle m_k\ |k\rangle$, where $|k\rangle$ donates the possible states (of valence, for example) of the measurement, $m_k$ is the value obtained from the measurement (e.g. valence) and $\langle k|\Psi\rangle$ denotes probability amplitude for the state $|k\rangle$. The summation in $k$ means the superposition of all the possible states. The expectation is given by $\langle\widehat{M}\rangle = \langle\Psi|\widehat{M}|\Psi\rangle = \sum_k |\langle k|\Psi\rangle|^2 m_k$. Note $|\langle k|\Psi\rangle|^2$ is the probability of state $|k\rangle$. The tricky bit is the operator of measurement $\widehat{M}$. It can be biased to the real part of $|\Psi\rangle$ (as reflection, for example), making the imaginary part "invisible" to the external observer (who uses this questionnaire).

The mid-section of Figure 7 shows an epistemic framework that handles imaginative or possible worlds. Epistemics of the virtual (Hoorn, 2012) states that following from the observation problem, the physical universe cannot be known and that observers merely have a *mental representation* of it. This representation they call 'Reality,' which is their own particular take on the world. Within that specific Reality, observers assign a truth value to information, which may range from 'true' through 'possible' to 'false.' Truth values are attributed according to beliefs that are relatively stable (e.g., Earth is flat and fixated. God does not gamble). The belief system is derived from culture, education, science, and religion. Truth values, however, may change when new information is encountered (e.g., Earth is round and rotates around its axis).

The imaginative part of an observer's ontology is classified as Fiction (Figure 7, right-hand side). These are the observations that range from 'possible' to 'false.' In obvious cases, they refer to what happens in motion pictures, play acting, and soap series but they might as well be the 'observation' of cosmic ripples, the conceptualization of tachyons or the quest for dark matter.

When new data do not fall in line with known concepts (e.g., Higgs bosons have far less mass than expected), thorough investigation commences into the particulars of the phenomenon, doing epistemic appraisals of how 'realistic' the observed features are. The skin texture of a robot, for example, may feel very realistic whereas the voice is unrealistically synthetic.

From this, the ontological status of incoming information is assessed. New data is deemed more or less true, falling within fiction or reality, having more or less realistic qualities. Like this, superpositions may occur of "I observe a halo of dust and gas [Reality] although in itself a black hole is invisible [unrealistic].”[7] Or: “I know that the multiverse is nonsense [Fiction] but it makes the Standard Model of particle physics more complete [realistic].”

The bottom segment of Figure 7 shows how the Bloch sphere representation of Reflective and Affective processing partly falls into the mental representation of Reality of the observer and partly into the aspect the observer imagines (i.e. the Fiction s/he upholds). With Figure 7, we posit that quantum physics at the basis of neurology explains observations of concurrency in cognitive emotion psychology in that following from the observer’s beliefs, the construct of the reality (Re) of one dimension in $\mathbb{R}^3$ physical space unavoidably incorporates an imagined (Im) dimension in the complex coordinate space $\mathbb{C}^3$.

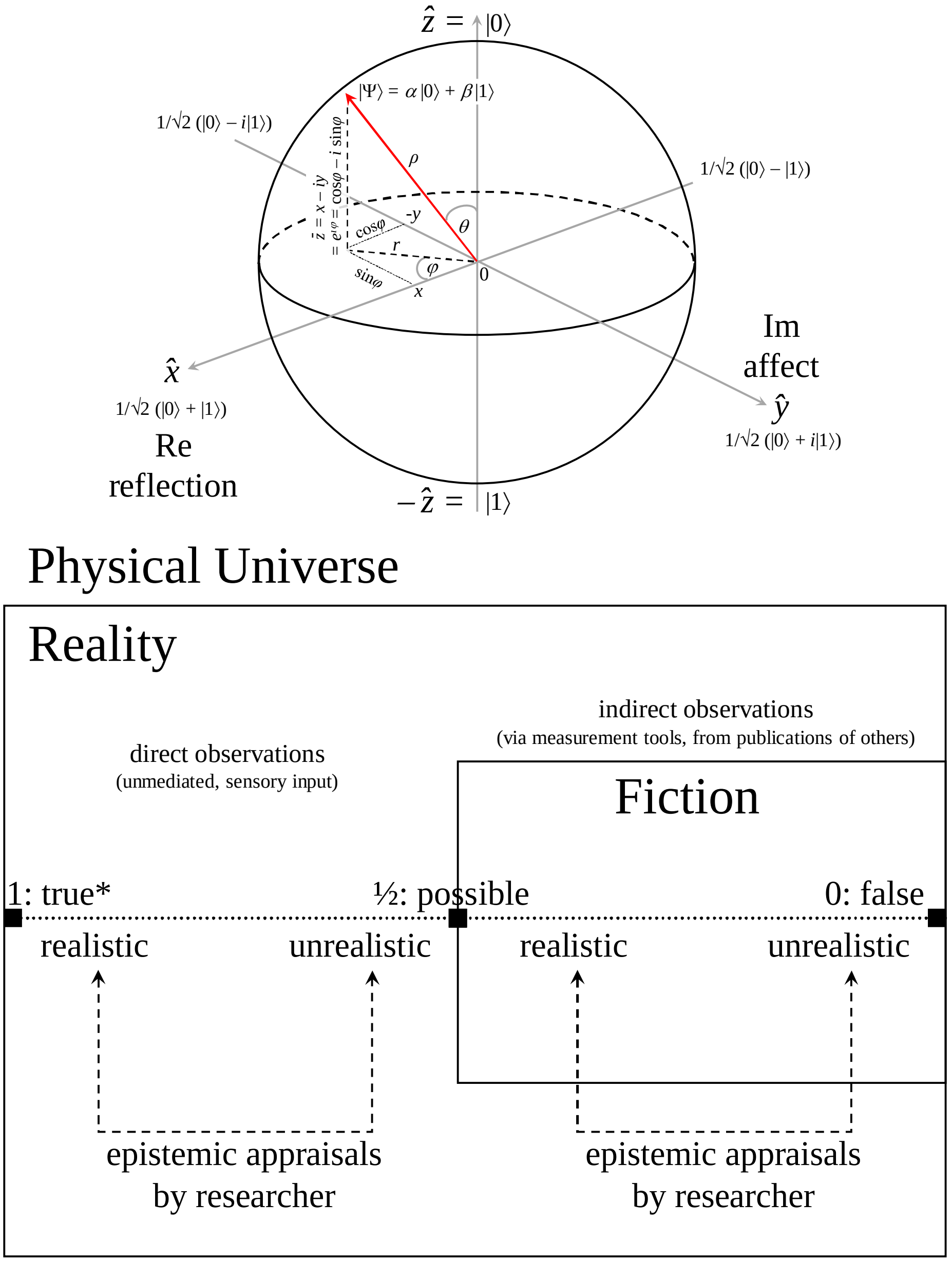

*Attribution of truth according to researcher’s belief system or world view (e.g., scientific, religious, or cultural)

[7] The second half of the statement is unrealistic because it follows from a fallacy of affirming the consequent.

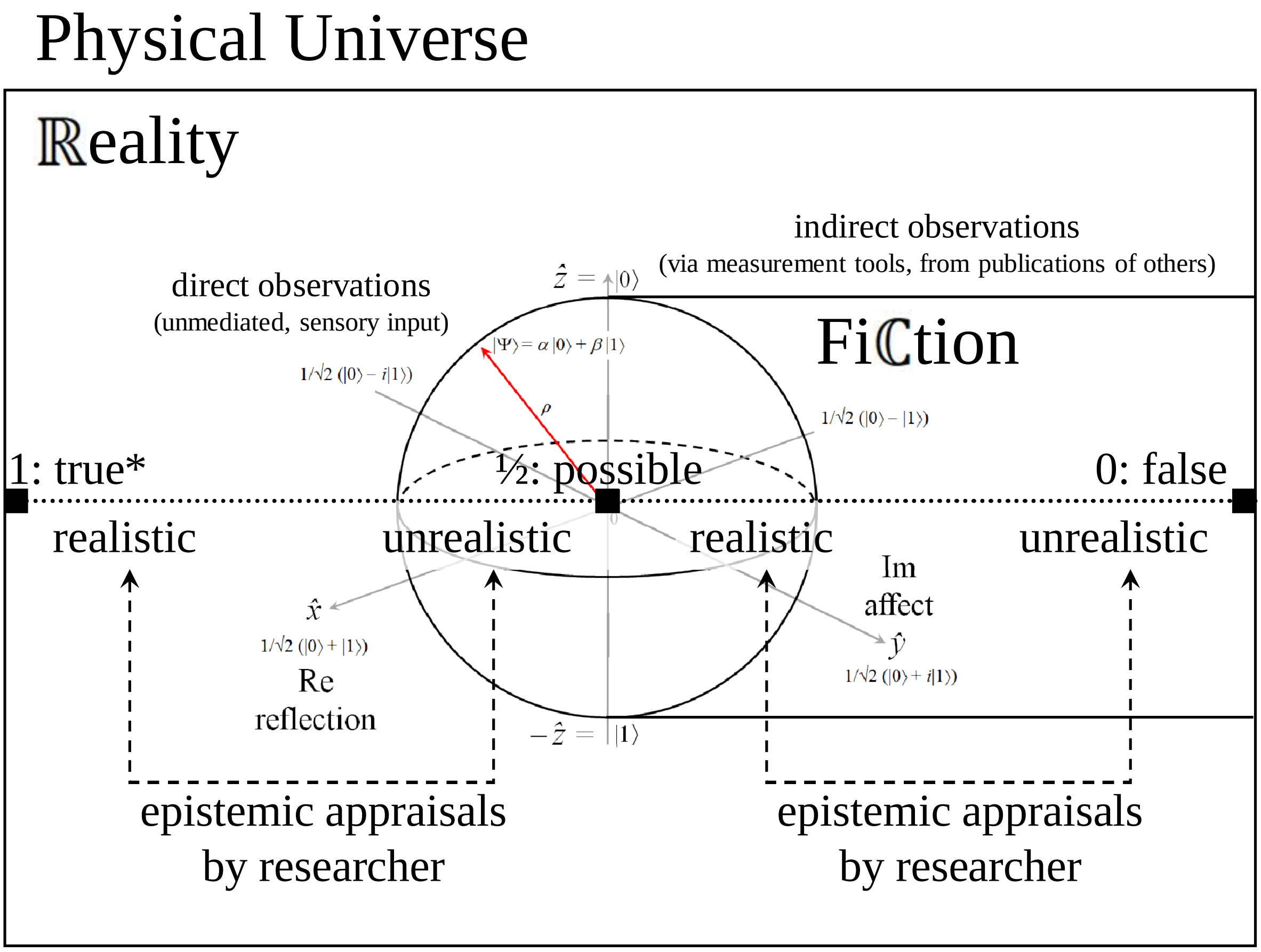

*Attribution of truth according to researcher's belief system or world view (e.g., scientific, religious, or cultural)

Figure 7:.Epistemics of the virtual for a Bloch representation of observed reflection and estimated affect.

With Figure 7 in mind, we now can assign a psychological meaning to each of the quantum-physical variables in the equation for the psychological state vector in (2). By the end of the section *Relevance, Valence | Reflection*, we defined $\alpha\,|0\rangle$ as the measure of relevance of affect (visible in the intensity of emotion) and $\beta\,|1\rangle$ as the relevance of a reflection (visible in the intensity of being in thought).[8] With this definition of $\alpha$ and $\beta$ in equation (2), $\cos(\theta/2)|0\rangle$ designates the Relevance of Affect, while $\sin(\theta/2)e^{i\varphi}|1\rangle$ relates to the Relevance of Reflection, where $\theta$ and $\varphi \in \mathbb{R}$ are part of (perceived) Reality (Figure 7) and $i \in \mathbb{C}$ is Fictitious (Figure 7). The result of this consideration is found in Figure 8.

[8] This is not 'depth' of thought, which relates to the number of brain structures employed by the cortex (Haller, 2016). Intensity here pertains to electric potential differences.

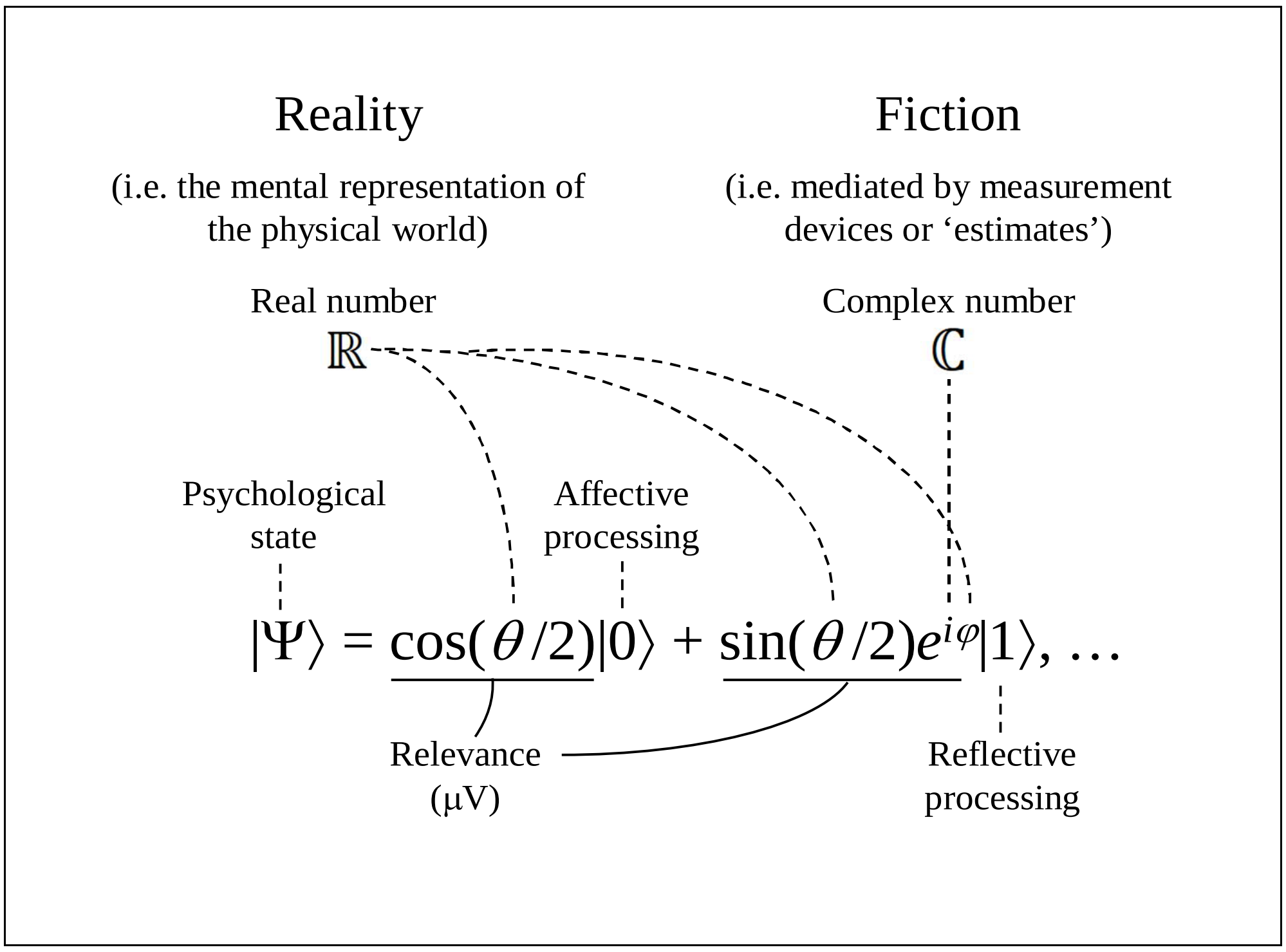

Figure 8: The psychological meaning of quantum-physical variables.

**Measurement against all odds: the anger-and-joystick**

The fundamental observation problem, which results in a lack of precision, does not invite to even want to measure an event let alone construct a new instrument, which is destined to be defective like all other. Nonetheless, multiple ways of measurement might show some consistency that is not an artefact of the device anymore (although admittedly still of human pattern recognition). Therefore, the next paragraphs unfold our future research plans of measuring affective and reflective processing, matching the Bloch-sphere view.

Imagine a transparent sphere of 3 meters in diameter with dashed laser beams, each in a different color, running over the *x*, *y*, and *z*-axis. The ends of the axes have labels that show the polarity of the scale: *x* (deep – shallow thought), *y* (positive – negative feelings), *z* (“I feel”– “I think”). The dashes indicate the intensity from the origin (weak) to the ends (strong).

The experimental trials require either more affective processing (e.g., evaluating disgusting pictures), more reflective processing (e.g., problem solving), or mixes of both (e.g., showing empathy). Inside the transparent sphere flies a mosquito-like micro air vehicle or a ‘mini drone of emotion’ if you will, handled by the participant after a training period. While processing the information offered on the experimental trials, participants navigate the Bloch sphere with an *anger-and-joy*stick that uses left and right to fly over the *x*-axis, forward (up) and back (down) to go over the *y*-axis, and rotating left (counter-clockwise) and right (clockwise) to move over the *z*-axis. To guarantee concurrency, on half the trials left hand use is for affective responses (*y*, *z*) and right hand use for reflective responses (*x*, *z*) and on the remaining trials the assignment to left and right hand is reversed. Rotating left and right to move the drone over *z* is making the choice (eigenvalue collapsing to $|0\rangle$ or $|1\rangle$).

A computer tracks the drone in 3D space and the flight path of the mosquito air-vehicle is compared to the series of points indicated by the state vector $|\Psi\rangle$ as predicted by a quantum computer that runs our (to be developed) Bloch-based Quantum-affective-computation software (Quaffection or $Qa^{-}$ for short).

While learning over trials and across participants, performance of the $Qa^{-}$ software is improved and so our model of human cognitive-affective processing becomes more accurate, which then may be installed on a quantum version of a robot brain server ($RBS^{|q\rangle}$) (Hoorn, 2018b). Then we may build a quantum application for a social robot (e.g., Hanson's Sophia or Chen Xiaoping's Jia Jia) and conduct a Turing test with real users, surveying whether users believe the robot's behavior is driven by a human operator or by our $Qa^{-}$ software.

## Conclusions/Discussion

Theoretically, we did not come across a stumbling block that forbids the grounding of psychological parallelism in quantum physical events. The oscillations of electrons, which supposedly transport the information, well may be driven by quantum dynamics so that the modulation of positive and negative emotions transpires from wave functions in the Amygdala structure that spread out to different numbers of brain resources summoned by the Neocortex.

Arguably, the regression-based path models of psychological functions hardly can deal with different areas of the brain that regulate different functions, flexibly, not discretely, while definitely working in collaboration (cf. Haller, 2016). With a quantum approach, we can account for brain regions that are strongly intertwined functionally, together arranging a sublime configuration of distributed processing. Moreover, with quantum dynamics, we go beyond the problematic correlations, which describe no causality and when time-based are meaningless altogether.

So how physical is it? We avow that ours is not a mere modeling exercise but describes actual physical events underlying psychological experience. We may be on the brink of reproducing in a computer what actually happens in the brain. The particle that carries the information is the electron. Each electron, then, spins inside a Bloch sphere, moving between $|0\rangle$ and $|1\rangle$ along the $\hat{z}$-axis. Thus, if bursts of electrons carry information of emotional relevance while $|0\rangle$ means 'process affectively' and $|1\rangle$ means 'do so reflectively,' many Bloch spheres may occur in which information-carrying electrons *literally* and *physically* rotate into the direction of the one or the other opposite polarity inside the Amygdala structure. Thus, one could jokingly maintain that e-motion indeed is electron motion!

Up till now we used fuzzy algorithms to model the way agencies build up affect for one another (e.g., Hoorn, 2008; Hoorn, Baier, Van Maanen, & Wester, submitted) but with quantum logics, the 50% membership of a feature in a set can now be represented as $1/\sqrt{2}\ |0\rangle + |1\rangle$ on the Bloch sphere. Even fuzzy logics is more conventional than quantum logics. First, fuzzy sets may represent the result of measurement but do not reflect what goes on in unobserved situations (Raghuvanshi & Perkowski, 2010). Second, the order in which one adds or multiplies is indifferent in fuzzy logics (ibid.). Qubits, however, hold more information on the psychological states before measurement and *do* produce different results based on the order of addition and multiplication. Third, fuzzy sets cannot represent entanglement (ibid.).

In their seminal paper, Raghuvanshi and Perkowski (2010) show a number of procedures to interrelate quantum states and fuzzy logics, which allows for a smooth transition from the one approach into the other. In this light, next goes an example of the usefulness of quantum modeling. Suppose someone buys a monitoring robot because of its usefulness and then suddenly that user's opinion changes once s/he finds out it has the company's spyware installed, then measurement = quantum bit1. Ergo, based on this one undesirable feature, the whole robot becomes undesirable despite its useful aspects (i.e. entanglement). Yet, if the observation of spyware remains absent (measurement = quantum bit0), then qubits do not collapse to distinct eigenvalues, maintaining a '(non-maximum) entanglement state' (Raghuvanshi & Perkowski, 2010).

To improve the interaction between artificial systems and their users, the current trend in social robotics is to make such systems not just intelligent but also emotionally sensitive. The research in affective computing stumbles upon fundamental issues of seriality in the computational hardware whereas the human brain fundamentally works differently. The physical make-up of the human brain is such that information is processed rationally and emotionally simultaneously. In the area of social robotics, this conundrum to conventional computers ensues emotionally clumsiness in a robot's behaviors, inhibiting the development of genuine partners for life for those who are socially isolated or deprived. Our proposal is to simulate the simultaneous cognitive-affective processing of information on a quantum computer to establish sentient machines that are emotionally intelligent.

To our knowledge, there are but two (!) precursors in this uncharted territory: Raghuvanshi and Perkowski (2010) and Yan et al. (2015). These are impressive first attempts. They model single emotions yet ignore the influence of rational processes. In the present paper, the affective process as such is modeled in unison with cognitive regulatory processes on the emotional outcomes (e.g., do not fight but discuss) and is connected to actual electromotive behaviors of subatomic particles. Our integration of quantum physics, neurology, emotion psychology, epistemology, and mathematics is unprecedented and unique. With it, we may build a quantum application for a social robot (e.g., Hanson's Sophia) and conduct a Turing test with real users, surveying whether users believe the robot's behavior is driven by a human operator or our quantum affective software. If successful, that research ushers in a quantum leap into the future.

## Conflicts of Interest

The authors declare that there is no conflict of interest regarding the publication of this paper.

## Funding Statement

This study was performed as part of the employment of the authors at The Hong Kong Polytechnic University, Vrije Universiteit Amsterdam, and Hong Kong Baptist University.

## Acknowledgments

We would like to acknowledge Ivy Shiming Huang for her advice on the Chinese Abstract.